\documentclass[letterpaper, 10 pt, conference]{ieeeconf}  

\IEEEoverridecommandlockouts                              

\usepackage{eqparbox}
\usepackage{multicol}
\usepackage{multirow}
\usepackage{diagbox}
\usepackage{slashbox}
\usepackage{amsmath}
\usepackage{algorithm}
\usepackage[noend]{algpseudocode}
\usepackage{array}
\usepackage{amssymb}
\usepackage{enumerate}
\usepackage{graphicx}
\usepackage{amsmath}
\usepackage{epstopdf}
\usepackage{url}
\usepackage{booktabs}
\usepackage{breqn}
\usepackage{afterpage}
\usepackage{cite}
\usepackage{subcaption}

\usepackage{caption}
\usepackage{tablefootnote}
\usepackage{threeparttable}
\usepackage{tikz}
\usepackage{lipsum}
\usepackage{multicol}
\usepackage{multirow}
\usepackage{hyperref}
\usepackage{url}

\usetikzlibrary{arrows,calc,positioning}
\tikzstyle{intt}=[draw,text centered,minimum size=6em,text width=5.25cm,text height=0.34cm]
\tikzstyle{intl}=[draw,text centered,minimum size=2em,text width=2.75cm,text height=0.34cm]
\tikzstyle{int}=[draw,minimum size=2.5em,text centered,text width=3.5cm]
\tikzstyle{intg}=[draw,minimum size=3em,text centered,text width=6.cm]
\tikzstyle{sum}=[draw,shape=circle,inner sep=2pt,text centered,node distance=3.5cm]
\tikzstyle{summ}=[drawshape=circle,inner sep=4pt,text centered,node distance=3.cm]
\def\eg{\emph{e.g., }}
\def\ie{\emph{i.e., }}
\def \ea#1{\eacitedA#1,,\end}
\def \eacitedA#1,#2,#3\end{#1~et~al\ifx,#2,\else .~\cite{#2}\fi}

\title{\LARGE \bf
Pedestrian Emergence Estimation and Occlusion-Aware Risk Assessment for Urban Autonomous Driving
}

\author{Mert~Ko{\c c}$^{1}$, Ekim Yurtsever$^{1}$, Keith Redmill$^{1}$, {\"U}mit~{\"O}zg{\"u}ner$^{1}$
\thanks{$^{1}$The Department of Electrical and Computer Engineering, 
            The Ohio State University, Columbus, Ohio, USA
            }%
\thanks{{\tt\small  \{koc.15, yurtsever.2, redmill.1, ozguner.1\}@osu.edu}}%
}

\begin{document}

\maketitle
\thispagestyle{empty}
\pagestyle{empty}

\begin{abstract}
Avoiding unseen or partially occluded vulnerable road users (VRUs) is a major challenge for fully autonomous driving in urban scenes.
However, occlusion-aware risk assessment systems have not been widely studied. Here, we propose a pedestrian emergence estimation and occlusion-aware risk assessment system for urban autonomous driving.
First, the proposed system utilizes available contextual information, such as visible cars and pedestrians, to estimate pedestrian emergence probabilities in occluded regions.
These probabilities are then used in a risk assessment framework, and incorporated into a longitudinal motion controller. 
The proposed controller is tested against several baseline controllers that recapitulate some commonly observed driving styles. 
The simulated test scenarios include randomly placed parked cars and pedestrians, most of whom are occluded from the ego vehicle's view and emerges randomly. 
The proposed controller outperformed the baselines in terms of safety and comfort measures.

\end{abstract}

\section{INTRODUCTION}
612,500 pedestrians were killed in 2013 by road traffic injuries, which was the number one cause of death among the age group 15-29 \cite{world2015global}.
Fully automated driving systems are seen as possible remedies for reducing road traffic fatalities due to the fact that they do not possess the fundamental issues of human drivers, such as failure to comply with the rules, lack of attention while driving, etc. 
Furthermore, decision-making for automated driving systems (ADS) is a challenging area that plays a key role in fully automated systems. Especially, developing intelligent systems taking precautious actions for objects that are currently unobservable but interacting with the ego vehicle in the future has attracted much attention recently.  
%
Currently, only up to Level~3~systems~\cite{sae_autonomylvl} are available in the market 
\cite{nhtsa_2021}. 

The motion prediction and risk assessment are vital in taking precautious actions. Lefevre et al.~\cite{lefevre2014survey} categorized the motion prediction and risk assessment methods into four categories, and 
stated that despite being computationally more demanding, interaction-aware methods are more reliable than other methods.
As an alternative to the interaction-aware methods, occlusion-aware \cite{shalev2016safe} risk assessment methods have been proposed. 
Those alternatives vary from solutions based on partially/mixed observable Markov decision processes (POMDP/MOMDP)
\cite{bandyopadhyay2013intention,brechtel2014probabilistic,bouton2018scalable,schratter2019pedestrian}
, to solutions based on set-based methods \cite{hoermann2017entering,lee2017collision,orzechowski2018tackling,tacs2018limited,yu2019occlusion,naumann2019safe}. Set-based methods and exploitation of behavior of other participants in a rule-based fashion were used \cite{althoff2014online,magdici2016fail,kousik2017safe,koschi2018set,kapania2019hybrid}. 
Althoff et al.~\cite{althoff2014online} improved the reachability analysis to obtain PID controllers and implemented their method on a real car.

There are currently two shortcomings of available methods in the literature. First, little attention has been paid to using visible information and prior knowledge to predict pedestrian emergence out of occluded areas. Second, human driver performance under occlusion and limited visibility conditions have been mostly neglected.
%

%

\begin{figure}[t]
\centering
\vspace{0.25cm}
\includegraphics[width=1.0\linewidth]{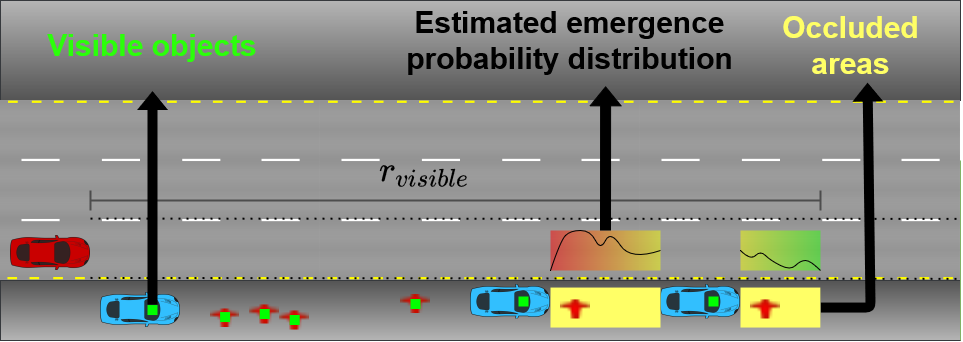}
\caption{An example of environmental cues and how a human driver would easily decide a similar scheme for emergence probability}
\label{fig:scene_example}
\vspace{-0.3cm}
\end{figure}

This paper introduces a novel occlusion-aware risk assessment system for ADSs.
The proposed method can estimate pedestrian emergence probability from occluded areas and adjust its driving policy accordingly. 
Our method's overview is demonstrated in fig.~\ref{fig:scene_example}, where the red ego vehicle uses information such as visible pedestrians and parked cars to assess the probability of emerging pedestrians to derive an optimal driving policy.

%
%
%
Our main contributions can be summarized as follows:
 \begin{itemize}
    \item Using contextual information for estimating pedestrian emergence from occluded areas
    \item Employing the estimated emergence probabilities in an occlusion-aware risk assessment framework
    \item Incorporating the proposed occlusion-aware risk assessment framework into 
    a longitudinal vehicle controller for realizing comfortable and safe driving
 \end{itemize}

The paper continues as follows: Section~\ref{sec:related} discusses previous work on the given subject. Section~\ref{sec:method} explains the framework of the current proposal in detail. Section~\ref{sec:evaluation} introduces 3 baseline controllers to compare against the proposed method and provide details about the simulation environment. Section~\ref{sec:results} evaluates the proposed method and demonstrates that the proposed method's performance is significantly better over several metrics quantitatively. Section~\ref{sec:conclusion} concludes and discusses possible future directions.

 \section{RELATED WORK}
 \label{sec:related}



\textbf{Driving policies without occlusion awareness.} 
\ea{Magdici,magdici2016fail}
generated so-called fail-safe optimal trajectories, some of which were designated as ``emergency maneuvers" so as to stop the ego vehicle without any collision; furthermore, the ``emergency-maneuvers" in
\cite{magdici2016fail}
generated by predicting all possible trajectories of other traffic participants within the given time horizon.
\ea{Kousik,kousik2017safe}
utilized the duality of low-fidelity and high-fidelity models to generate safe trajectories in arbitrary environments in real-time; safe trajectories were generated by taking into account the mismatch between the models.
\ea{Koschi,koschi2018set}
used set-based predictions, contextual information, and traffic rules. 
However, only visible pedestrians were considered in \cite{koschi2018set}. 
\ea{Kapania,kapania2019hybrid}
first utilized the gap acceptance behavior of pedestrians and deterministic limitations of vehicles. The gap acceptance behavior and the limitations were used by an FSM controller. The controller yielded whenever possible, or chose the action to avoid collisions; furthermore, the controller exploited the gap acceptance behavior in the different modes, such as aggressive and conservative. However, none of those works considered occlusions and invisible traffic participants.

On the other hand, occlusion-aware risk assessment methods incorporated obstructed visibility information. These methods can be categorized into two: the methods based on POMDP/MOMDP representation and the methods based on reachability analysis.

\textbf{POMDP/MOMDP representation.} These methods use probabilistic models and Bayesian filtering \cite{bandyopadhyay2013intention,brechtel2014probabilistic,bouton2018scalable,schratter2019pedestrian}. These methods could estimate the uncertainties to incorporate the observation along with reward models. Then, the estimation was used to extract a sub-optimal driving policy.
\ea{Bouton,bouton2018scalable}
extended the work of \cite{bouton2018utility}~and~\cite{russell2003q} on utility decomposition to propose a scalable decision-making method under sensor occlusions; the proposal could handle multiple road users. However, the action and state resolution were very low on the account of avoiding exponentially growing computational demand. 
\ea{Schratter,schratter2019pedestrian}
achieved safe control for occluded pedestrians with POMDP representation and the help of Automatic Emergency Braking (AEB) systems. However, that method suffered from the low resolution of state-space, and it could handle one pedestrian.
In general, methods relying on POMDP representation and RL have suffered from real-time inapplicability. In other words, those methods are computationally demanding; consequently, they are impractical for real-life implementations as the execution time is rather long in situations that require immediate actions. Secondly, due to lack of resolution in action and state-space, the driving policy may result in an uncomfortable driving experience for passengers. 

\textbf{Reachability analysis.} These methods determine the risky situations in advance by over-approximating the future sets of other traffic participants using reachability analysis \cite{hoermann2017entering,lee2017collision,orzechowski2018tackling,tacs2018limited,yu2019occlusion,naumann2019safe}. 
\ea{Hoerman,hoermann2017entering}
used dynamic grid maps in three distinct dynamic grid map layers, namely object-based, object-free, and unobservable region layer for risk assessment and collision-avoidance; any cell was considered as occupied had the cell in any of the three layers detected as occupied. 
\ea{Orzechowski,orzechowski2018tackling}
expanded
\ea{Althoff}'s
reachability analysis so as to also incorporate occlusions and unobservable traffic users. Furthermore, Orzechowski et al. stated that Ta{\c s}~and~Stiller~\cite{tacs2018limited} prioritized safety, did not consider comfort. 
Naumann et al.~\cite{naumann2019safe} considered the worst-case presumptions for critically unseen traffic users and claimed to ensure ``provably safe but not over-cautious actions". Furthermore, most of these methods adopted the notion of blame from \cite{shalev2016safe}, which defined safety as choosing trajectories that do not cause a collision. 
However, those trajectories may still end in a collision due to erroneous actions by other traffic participants; moreover, neglecting to compensate for other's faults is a major flaw that may result in morally wrong situations.
Those methods assumed the worst-case scenarios in some subset of possible actions by other traffic participants; specifically, for those which considered the pedestrians in over-approximated sets. However, a proof of safety by constraining physical capabilities of other traffic participants is ill-defined, in that there may exist some cases that those constraints are either too conservative or not sufficiently representative.

None of the methods has taken into account the fact that the contextual information could be incorporated into the prior knowledge; hence, the information can be utilized to assess the risk for occluded unseen pedestrians. In addition, the literature is deficient in formally defining the limitations of vehicles in terms of safety, and comfort. For example, the distance to stop without a compromise on safety and the distance to stop without a compromise on passengers' comfort change with different weather conditions. Consequently, we believe that considering the limits of safety and comfort for different conditions will improve the robustness and capabilities of ADS.



 \section{METHODOLOGY}
 \label{sec:method}
\begin{figure*}[t]
    \centering
    \includegraphics[width=\textwidth]{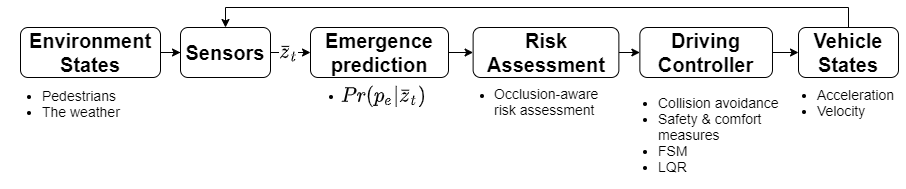}
    \caption{Overview of the proposed framework}
    \label{fig:proposed_flow}
\end{figure*}
Here, we propose a novel occluded emerging pedestrian distribution estimation and risk assessment method. 
The framework includes three parts: (1) estimating pedestrian emergence from occlusions; (2) risk assessment; and (3)~the controller.
The overview of the proposed method is demonstrated in fig.~\ref{fig:proposed_flow}. 
A simulation environment is created using Python programming. 
The parked cars are placed randomly on the two sides of the road to create the occlusions. 
The occlusions and visible objects to the ego vehicle are calculated using a simple visibility polygon algorithm. 
For this task, we assume that the ego vehicle has sensors that can identify each object and locate the visible objects within the visible range $r_{visible}$ and the viewing angle. Nevertheless, our proposal can be compatible with any sensors that can detect, \ie can label an object as in ``a vehicle" or ``a pedestrian", and locate the objects.
To generate realistic scenarios, pedestrians are modeled as point-mass objects with instant velocity change because the time passed until a pedestrian accelerates to its walking velocity is negligible.
We investigated the empirical data of the pedestrian velocity from \cite{saberi_aghabayk_sobhani_2015}. Accordingly, we used a Gaussian distribution with $\mu_{v\_ped} = 1.5 m/s$ and  $\sigma_{v\_ped} = 0.6 m/s$. 
In addition, since not all the pedestrians that a driver sees on the sidewalk will cross the street, we generated some pedestrians that would not cross the street. 
\subsection{Estimating Pedestrian Emergence from Occlusions}
The goal is to find the distribution of the emerging pedestrians from occlusions. Our solution is to use contextual information such as the presence of the parked cars, crosswalks, and visible pedestrians. 
As we set forth previously, the POMDP representations and Bayesian filtering are computationally demanding. Therefore, we suggest a solution that generates a posterior belief without explicit Bayesian filtering.
Specifically, to reduce the computational demand of the estimation of the posterior distribution, a collection of piecewise weighted sigmoid functions was utilized. 
%
%
%
%
Then, the probability approximation becomes: 
%
\begin{gather}
    \textrm{Pr}(p_e|\bar{z}_t) \simeq \frac{1}{1 + e^{-\bar{w}^T \bar{z}_t}} \label{eq:riskfunc}\\
    \bar{z}_t = [1, n_1, n_2, d_1, d_2, d_3]^T\label{eq:obs}
\end{gather}
where $\textrm{Pr}(p_e|\bar{z}_t)$ is the pedestrian emergence from occlusion probability, $p_e$ is the pedestrian emergence event, and $\bar{z}_t$ is the observation vector; $n_1$ is the normalized density of parked cars, $n_2$ is the normalized density of visible pedestrians, $d_1$ is the normalized distance to the crosswalk, $d_2$ is the normalized distance to the closest parked car, and $d_3$ is the normalized distance to the closest visible pedestrian. 
Heuristics have chosen the weight vector of the observations in \eqref{eq:riskfunc}.
%
%
In the case of unobservability, $n_1$, $n_2$, $n_3$ is considered 0 whereas $d_1$, $d_2$, $d_3$ considered 1 which is the normalized value of $r_{visible}$. 

\subsection{Risk Assessment}
\label{sec:risk_assessment}
Risk assessment is the most crucial part and the main contribution of this work.
Passengers can feel both the force and the change in the force exerted on their bodies. The change in force is widely known as \textit{jerk}. Furthermore, both acceleration and jerk are perceived omnidirectionally by a human body. Therefore, a $a_{comfort}$, and a $j_{comfort}$ value can be defined under the assumption that values between $[-a_{comfort}, a_{comfort}]$ for acceleration, and values between $[-j_{comfort}, j_{comfort}]$ for jerk are defined as \textit{comfortable}.

Due to physical limitations, before reaching a steady acceleration of choice, the acceleration's magnitude rises linearly for a ramp time $t_{ramp}$.
Then, the ramp time from the magnitude of $0$ to $a_{comfort}$ is denoted by $t_{ramp,comfort}$ whereas the ramp time from the magnitude of $0$ to $a_{max}$ is denoted by $t_{ramp,min}$. 
Also, the distance traveled before stopping by the ego vehicle, with the magnitude of deceleration decided to be $a_{comfort}$ and the ramp time decided to be $t_{ramp,comfort}$, is denoted by $d_{stop,comfort}$ whereas the distance traveled with the magnitude of deceleration decided to be $a_{max}$ and the ramp time physically possible being $t_{ramp,min}$ is denoted by $d_{stop,min}$
Note that, $t_{ramp,min}$, $a_{max}$, and $d_{min}$ are physical limitations and constant for the ego vehicle whereas $t_{ramp,comfort}$, $a_{comfort}$ and $d_{stop,comfort}$ are variables. 
Moreover, the minimum distance using the comfortable values is denoted by $d_{stop,comfort,min}$.
Using the notation, the ego vehicle can have imaginary zones; here, it is defined as \textit{risk zones}. 
Assuming that $d_{stop,min} \leq d_{stop,comfort,min}$, the so-called risk zones:
\begin{itemize}
    \item \emph{danger zone} spans $[0, d_{stop\_min}]$
    \item \emph{discomfort zone} spans $[d_{stop\_min}, d_{stop\_comfort}]$
    \item \emph{safety zone} spans $[d_{stop\_comfort}, r_{visible}]$ meters ahead, away from the ego vehicle.
\end{itemize} 
\begin{algorithm}[t]
\caption{The proposed algorithm (Part 1)}
\label{alg:prposedcontroller}
\begin{algorithmic}[1]
\Function{proposedController}{}
\State $a_{max} \gets \mu_{road} *  g$
\State $s_{t+1} \gets s_{\textrm{normal drive}}$
\If{A visible pedestrian is to be inside the path}
    \If{$TTC < TTC_{\textrm{stop}}$}
        \State $s_{t+1} \gets s_{\textrm{emergency}}$
    \Else
        \State $s_{t+1} \gets s_{\textrm{yielding}}$
    \EndIf
\Else
    \State $current\_state \gets danger$
    \State $max\_risk \gets 0$, $current\_risk \gets 0$
    \Repeat
        \State $\#$ Update current \emph{risk zone}
        \If{$d > d_{stop\_min}$}
            \State $current\_state \gets discomfort$
            \State $max\_risk \gets 0$
        \EndIf
        \State $\#$ Check the neighboring visible cues
        \State $current\_risk \gets $ from \eqref{eq:riskfunc}
        \If{$max\_risk < current\_risk$}
            \State $max\_risk \gets current\_risk$
        \EndIf
        \If{$current\_state = danger$}
            \State Set $l_{cautious}$, $l_{steady}$, $a_{limit}$ and $j_{limit}$
            \If{$max\_risk > l_{steady}$}
                \State $s_{t+1} \gets s_{\textrm{steady drive}}$
            \ElsIf{$max\_risk > l_{cautious}$}
                \State $s_{t+1} \gets s_{\textrm{cautious drive}}$
            \EndIf
        \ElsIf{$current\_state = discomfort$}
            \State Set $l_{discomfort}$, $l_{steady}$, $a_{limit}$ and $j_{limit}$
            \If{$max\_risk > l_{steady}$}
                \State $s_{t+1} \gets s_{\textrm{steady drive}}$
            \ElsIf{$max\_risk > l_{cautious}$}
                \State $s_{t+1} \gets s_{\textrm{cautious drive}}$
            \EndIf
        \EndIf
        \State $d \gets d + \Delta d$
    \Until{$d \geq d_{stop\_comfort}$}
\EndIf
\State \textbf{return} $s_{t+1}$
\EndFunction
\end{algorithmic}
\end{algorithm}
\noindent
\subsection{Driving Policy}
The proposed occlusion-aware risk assessment framework is incorporated into a safe and robust driving policy.
The driving policy is given in algorithm~\ref{alg:prposedcontroller}.



\textbf{Longitudinal control.}
A controller is required to achieve the necessary control actions.
We use a modified, LQR-based control strategy based on a point-mass discrete-time vehicle dynamics model given by:
\begin{align}
    \bar{x}_{crs,k+1} &= v_{k+1} = v_k + \Delta t * a_k \label{eq:discrete_vel}\\
    d_{k+1} &= d_k - \Delta t * v_k \label{eq:discrete_dist}
\end{align}
where $v_k$ is the velocity of the ego vehicle, $a_k$ is the acceleration value of the ego vehicle  $d_k$, in \eqref{eq:discrete_dist}, is the lateral distance to the imaginary line tangent to the close side of the visible pedestrian to the vehicle, and also perpendicular to the ego vehicle's direction. Here, we used $k$ to distinguish discrete-time representation from continuous-time representation. 
Henceforth, we replace $k$ with $t$.
Using the vector notation again with the discrete time equations:
\begin{gather}
    \bar{x}_{yld,t+1} = \begin{bmatrix}
    d_{t+1}\\
    v_{t+1}
    \end{bmatrix} =\begin{bmatrix}
    1 & -\Delta t\\
    0 & 1
    \end{bmatrix} \begin{bmatrix}
    d_t\\
    v_t
    \end{bmatrix} + \begin{bmatrix}
    0\\
    \Delta t
    \end{bmatrix} \bar{u}_t\\
    \bar{u}_t = a_t
\end{gather}
When there is no visible pedestrian to be yielded, the state-space is represented by $\bar{x}_{crs,t}$.
By contrast, when there is at least one visible pedestrian, the state-space is represented by $\bar{x}_{yld,t}$; 
the distance between the closest pedestrian and the vehicle is one of the states. 
Then, the traditional LQR optimization scheme uses the quadratic cost function to optimize the control action. The cost function is defined as follows:
\begin{gather}
    c_t = (\bar{x}_t - \bar{x}_{ref})^T Q (\bar{x}_t - \bar{x}_{ref}) + \bar{u}_t^T R \bar{u}_t\\
    J = \sum _{t=0}^{T-1} c_t \label{eq:cum_cost}\\
    \bar{u}_t = -K (\bar{x}_t - \bar{x}_{ref})
\end{gather}
where $K$ is the Kalman gain, and the optimal control is obtained by minimizing the cumulative cost $J$ provided that appropriate $Q$ and $R$ matrices are selected. 

However, the traditional LQR is an unconstrained-optimization scheme. Consequently, this proposal with the traditional LQR will generate unrealistic and high-frequency control actions under noise in the measurements or the system.
Those high-frequency modes of the action can be very uncomfortable and dangerous for the passengers.
The equations of the LQR scheme can be revised with a small modification to limit the jerk:
\begin{align}
    a_{t+1} &= a_t + \Delta t * j_t\\
    \begin{bmatrix}
    d_{t+1}\\
    v_{t+1}\\
    a_t
    \end{bmatrix} &= \begin{bmatrix}
    1 & -\Delta t & 0\\
    0 & 1 & \Delta t\\
    0 & 0 & 1
    \end{bmatrix} \begin{bmatrix}
    d_{t}\\
    v_{t}\\
    a_{t-1}
    \end{bmatrix} + \begin{bmatrix}
    0\\
    j_1 * \Delta t^2\\
    j_1 * \Delta t
    \end{bmatrix} j_t\label{eq:ssyielding}\\
    \begin{bmatrix}
    v_{t+1}\\
    a_t
    \end{bmatrix} &= \begin{bmatrix}
    1 & \Delta t\\
    0 & 1
    \end{bmatrix} \begin{bmatrix}
    v_{t}\\
    a_{t-1}
    \end{bmatrix} + \begin{bmatrix}
    j_2 * \Delta t^2\\
    j_2 * \Delta t
    \end{bmatrix} j_t \label{eq:sscruise}
\end{align}
where the state-space for yielding with limited jerk is described in \eqref{eq:ssyielding}, and the state-space for cruising with limited jerk is described in \eqref{eq:sscruise}. 
Since the jerk of a vehicle is not a realizable control input, the LQR controller's implementation with the modification will be similar; the controller will actuate the control input $a_t$ obtained from \eqref{eq:ssyielding} or \eqref{eq:sscruise}. 
$j_1$ is the maximum allowed jerk while yielding, whereas $j_2$ is the maximum allowed jerk while cruising as the ego vehicle must have more agility while yielding. 
Therefore, $j_1$ is chosen as $2m/s^3$, the maximum jerk for aggressive driving; by contrast, $j_2$ is chosen as $0.9m/s^2$ the maximum jerk for normal driving \cite{bae2020self}.

Finally, we have chosen appropriate $Q$ and $R$ matrices for both the \emph{cruising} and \emph{yielding} and computed the resulting full-state-feedback coefficients $K$ in MATLAB--R2019a as follows:
\begin{align*}
    Q_{crs} &= \begin{bmatrix}
    1000 & 0\\
    0 & 1\end{bmatrix}
    & R_{crs} &= \begin{bmatrix}
    1000\end{bmatrix}\\
    Q_{yld} &= \begin{bmatrix}
    5 & 0 & 0\\
    0 & 100 & 0\\
    0 & 0 & 0.1
    \end{bmatrix}
    & R_{yld} &= \begin{bmatrix}
    1500
    \end{bmatrix}
\end{align*}
\begin{gather*}
    K_{crs} = \begin{bmatrix}
    0.9047 & 0.9074\end{bmatrix}\\
    K_{yld} = \begin{bmatrix}
    -0.0532 & 0.3139 & 0.3792
    \end{bmatrix}
\end{gather*}
\textbf{Collision Avoidance.}
After calculating the span of so-called risk zones, assessing the current risk, it is also necessary to determine the collision risk.
The quantity measurements, \ie the normalized density are discrete; therefore, discontinuities in the function appear at the critical instance of observing or losing sight of a parked car or a pedestrian. 
Consequently, we use thresholds to divide the function into regions.
for a robust control architecture.

\begin{figure}[t]
    \centering
    \includegraphics[width=0.9\linewidth]{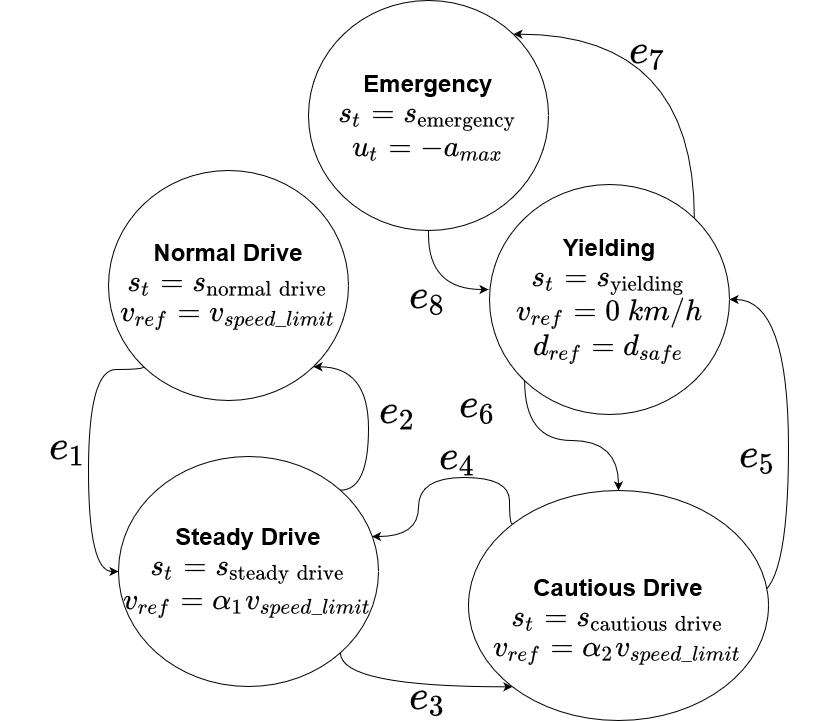}
    \caption{The proposed FSM to utilize the quantified risk}
    \label{fig:FSM}
\end{figure}

We designed an FSM controller with distinct states for each region in the function. As the risk decreases, the ego vehicle could be more encouraged to drive the speed limit while slowing down to a proportion of the speed limit in a risky situation. 
We named the FSM states for each region as follows:
\begin{itemize}
    \item \emph{Normal Drive}: The ego vehicle is given the reference velocity of the speed limit, $v_{speed\_limit}$,
    \item \emph{Steady Drive}: The ego vehicle is given the reference velocity of some proportion of the speed limit, $\alpha_1 v_{speed\_limit}$,
    \item \emph{Cautious Drive}: The ego vehicle is given the reference velocity of another proportion of the speed limit, $\alpha_2 v_{speed\_limit}$.
\end{itemize}
Also, we consider two additional crucial driving modes: yielding pedestrians when necessary and engaging in maximum braking to avoid collision under dangers. 
The resulting FSM architecture is demonstrated in fig.~\ref{fig:FSM}. Additionally, the state transition conditions of the designed controller are given in Table~\ref{tab:state_trans}. 
%
The interpretation of risk inside different \emph{risk zones} should be different because even a small risk of emergence in \emph{danger zone} should be treated with utmost care. In contrast, the same level of risk of emergence in \emph{discomfort zone} might not require similar alertness. 
Therefore, we assigned different threshold values for the different risk zones.

Combining all, the ego vehicle observes the environment from its sensors to predict the distribution of the emerging pedestrians from occlusions; this information is utilized, after risk assessment, differently per \emph{risk zone}, and this inference is going to be converted into a high-level command such as \emph{normal drive}, \emph{yielding}, etc.; finally, the controller is going to choose the appropriate control action, \ie acceleration, to reach the reference state within decided control limits.

The ego vehicle does not require to take precautious action against any pedestrians entering the expected path from inside the safety zone, \ie the algorithm's maximum look-ahead distance should be $d_{stop,comfort}$. 
Then, the future risk is computed by considering contextual information within a predefined window with the spatial resolution $\Delta d$.
The maximum risk per risk zone is compared against the thresholds to choose the appropriate deceleration/acceleration limit $a_{limit}$ and jerk limit $j_{limit}$ with the appropriate FSM state to reach the reference velocity $v_{ref}$. 
It is also crucial to determine the deceleration limit by the friction coefficient between the tires and the road. Here we assume that the ego vehicle can measure the friction coefficient for different weather conditions. 
\begin{table}[t]
\centering
\caption{State Transitions for the FSM in fig.~\ref{fig:FSM}}
\label{tab:state_trans}
\resizebox{0.8\linewidth}{!}{%
\begin{tabular}{@{}cc@{}}
\toprule
\textbf{\shortstack{State\\Transition}} & \textbf{\shortstack[c]{Explanation\\ \hspace{1cm}}}\\ \midrule
$e_1$  & $Pr(p_e|\bar{z}_t) > l_{steady}$ \\ 
$e_2$  & $Pr(p_e|\bar{z}_t) \leq l_{steady}$ \\ 
$e_3$  & $Pr(p_e|\bar{z}_t) > l_{cautious}$ \\ 
$e_4$  & $l_{steady} \leq Pr(p_e|\bar{z}_t) \leq l_{cautious}$ \\ 
$e_5$  & A visible pedestrian to be yielded\\ 
$e_6$  & No visible pedestrian to be yielded\\ 
$e_7$  & $TTC_{brake} \geq TTC_{emergency}$\\ 
$e_8$  & $TTC_{brake} < TTC_{emergency}$\\ 
\bottomrule
\end{tabular}
}
\end{table}
 \section{EXPERIMENTS}
 \label{sec:evaluation}
 In order to evaluate the proposal, a straight road is chosen. 
The road is 96 meters long, and it has three lanes whose width is 3 meters. The scenarios are divided into three categories based on the crowdedness; In \emph{suburban scenarios (sc1)}, there are one or two pedestrian, one or two parked cars and a crosswalk; in \emph{mildly crowded urban scenarios (sc2)}, there are multiple parked cars, multiple pedestrians and a crosswalk; in \emph{very crowded urban scenarios(sc3)}, the parking slots are full, and there are multiple pedestrians and a crosswalk.
This way it is going to be possible to observe the strengths and weaknesses of the proposal in various possible scenarios. 
We compare the proposal with the 3 baselines that represents a driving aspect. 
\begin{itemize}
    \item \emph{Baseline1 (B1)}: occlusion-unaware, and drives the legal speed limit of the road (30$km/h$). It yields to the pedestrians that are on the road will likely enter its expected path.
    \item \emph{Baseline2 (B2)}: occlusion-unaware, and drives the two-third of the legal speed limit of the road (20$km/h$). It yields to the pedestrians that are on the road will likely enter its expected path.
    \item \emph{Baseline3 (B3)}: occlusion-unaware, and drives the one-third of the legal speed limit of the road (10$km/h$) only if it observes a crosswalk which closer than a specific distance; otherwise, it drives the speed limit. It yields to the pedestrians that are on the road will likely enter its expected path.
\end{itemize}

\subsection{Metrics}
\label{sec:metrics}
%
An extensive survey on important metrics to determine the driving quality of AVs has been made by \cite{jahangirova2021quality}.
Combining the metrics from \cite{jahangirova2021quality} with additional safety metrics, we have decided to use the following metrics:
\begin{itemize}
    \item \emph{mt1}: The total number (successful/unsuccessful) of yields
    \item \emph{mt2}: Deceleration (mean, std)
    \item \emph{mt3}: The total number of successful finishes
    \item \emph{mt4}: Time of emergency braking (mean, std)
\end{itemize}
\begin{figure}[t]
    \centering
    \includegraphics[width=0.8\linewidth]{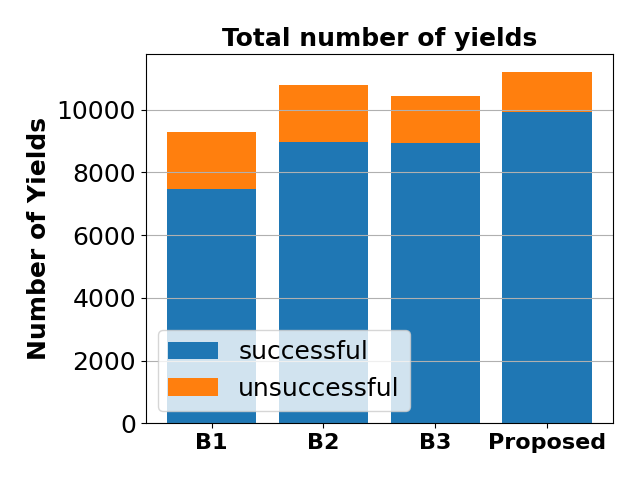}
    \caption{Total successful and unsuccessful yields (\emph{mt1}) of different controllers in \emph{mildly crowded urban scenarios} (\emph{sc2})}
    \label{fig:res_totalyields}
\end{figure}
\begin{table}[t]
\centering
\caption{Overall Performance of Proposed Controller and Baselines over 1000 episodes}
\label{tab:results}
\begin{threeparttable}
\begin{tabular}{@{}lcccr@{}}
\toprule
\textbf{Metrics} & \textbf{\emph{B1}} & \textbf{\emph{B2}} & \textbf{\emph{B3}} & \textbf{Proposed}\\ \midrule
\emph{mt1}
(\emph{sc1}) & (817/94) & (934/24) & (912/33) & \textbf{(937/23)}\\
\emph{mt1} (\emph{sc2}) & (7482/1811) & (8980/1821) & (8929/1496) & \textbf{(9911/1294)}\\
\emph{mt1} (\emph{sc3}) & (8012/1754) & (9645/1875) & (9449/1535) & \textbf{(10310/1404)}\\
\midrule
\emph{mt2}
(\emph{sc1}) & (-2.90, 1.55) & (-1.12, 0.99) & (-1.42, 1.23) & \textbf{(-1.07, 0.95)}\\
\emph{mt2} (\emph{sc2}) & (-2.13, 1.78) & (-1.31, 1.30) & (-1.05, 1.29) & \textbf{(-0.79, 0.90)}\\
\emph{mt2} (\emph{sc3}) & (-1.99, 1.75) & (-1.29, 1.28) & (-1.02, 1.27) & \textbf{(-0.80, 0.91)}\\
\midrule
\emph{mt3}
(\emph{sc1}) & 898 & 992 & 966 & \textbf{996}\\
\emph{mt3} (\emph{sc2}) & 652 & 916 & 856 & \textbf{986}\\
\emph{mt3} (\emph{sc3}) & 664 & 956 & 870 & \textbf{988}\\
\midrule
\emph{mt4}
(\emph{sc1}) & (0.27, 0.35) & (0.08, 0.18) & (0.09, 0.22) & \textbf{(0.07, 0.16)}\\
\emph{mt4} (\emph{sc2}) & (0.63, 0.57) & (0.32, 0.38) & (0.30, 0.38) & \textbf{(0.11, 0.18)}\\
\emph{mt4} (\emph{sc3}) &  (0.63, 0.60) & (0.30, 0.36) & (0.30, 0.39) & \textbf{(0.12, 0.19)}\\
\bottomrule
\end{tabular}
\end{threeparttable}
\end{table}
\begin{figure}[t]
    \centering
    \includegraphics[width=0.8\linewidth]{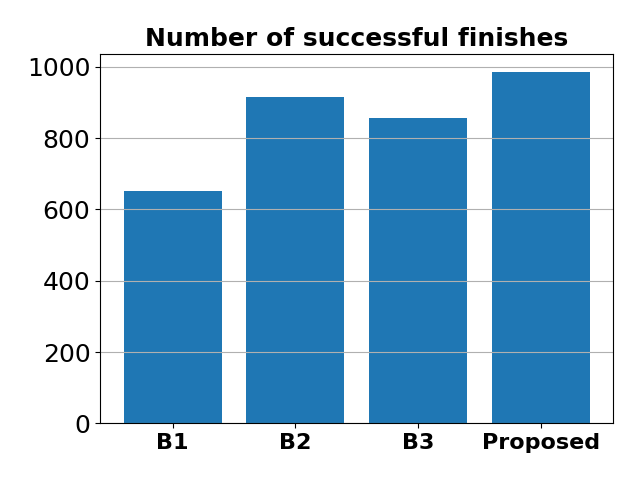}
    \caption{The total number of successful finishes (\emph{mt3}) of different controllers in \emph{mildly crowded urban scenarios} (\emph{sc2})}
    \label{fig:res_successfinish}
\end{figure}

 \section{RESULTS}
 \label{sec:results}
\begin{figure}[t]
    \centering
    \includegraphics[width=0.8\linewidth]{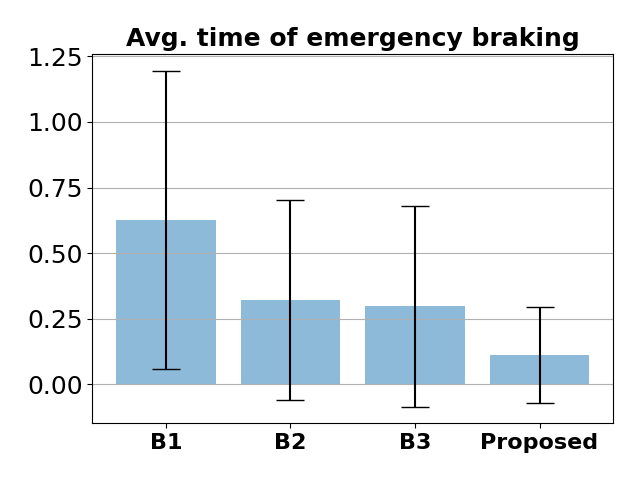}
    \caption{Average time of emergency braking (\emph{mt4}) of different controllers in \emph{mildly crowded urban scenarios} (\emph{sc2})}
    \label{fig:res_ebrake}
\end{figure}

 The results of the simulation, 1000 episodes per scenario, is demonstrated in Table~\ref{tab:results}.
In terms of comfort, the proposed method clearly outperforms in terms its yielding capabilities (\emph{mt1}) even though all baseline methods have the exact same FSM policy for yielding. Specifically, the proposed method outperforms the baselines in \emph{sc2}, by $32.46\%$, $10.37\%$, $11.00\%$ respectively. This clearly shows that, the proposed method is pedestrian friendly, in that without resorting to emergency braking, it yields to as many pedestrians as possible.
In terms of safety, outperformance of the proposed method is clearly visible for the number of successful finishes (\emph{mt3}) out of 1000 episodes per scenario where the proposed method outperforms the baselines in \emph{sc2} by $51.23\%$, $7.64\%$, $15.19\%$ respectively (performance in \emph{sc2} is demonstrated in fig.~\ref{fig:res_successfinish}). 
This metric is crucial as it is directly related to collision risk of the method.
Furthermore, the proposed method also outperforms all other baselines both in the \emph{average deceleration} (\emph{mt2}) and the \emph{average time of emergency braking} (\emph{mt4}) which could be interpreted as an indication of how successfully a method anticipates the incoming risk such that it resorts to minimal deceleration value and emergency braking (performance in \emph{sc2} is demonstrated in fig.~\ref{fig:res_ebrake}). One important remark is that in fig.~\ref{fig:res_ebrake} total number of yields per controller is different due to the fact that this metric is only available for successful path completions. In other words, an aggressive driving style may end up in a collision which in return means that the successful yields in the episode will not be considered. 

One limitation is due to the simplification of the risk assessment with another function. The weights of the function that assesses the risk are determined heuristically, and the optimality of resulting controller have not proven. Therefore, there may exist a better weight vector, or a better representation of the risk.

Another limitation the simplification such as assessing the risk on a straight road, assuming that the vehicle is a point-mass object might limit the performance of the proposed method. Although, we have considered several important and realistic phenomena, \eg the delay in actuating the control actions and the delay in sensing objects, the aforementioned assumptions might still deviate the implementation results from the simulated ones.
 \section{CONCLUSION}
 \label{sec:conclusion}
 This paper proposed a probabilistic risk assessment and collision avoidance method for emerging pedestrians from occlusions and demonstrates a possible proof-of-concept for the proposal. The proposal was compared against several baselines. 
The method  was evaluated against the baselines in three different scenarios, 1000 episodes with randomized initial conditions per scenario type, in the simulation environment in Python built from scratch. 
The method outperformed these baselines in the predefined metrics. 
Since the proposal does not rely on accurate map data or accurate and precise localization to achieve occlusion-aware vehicle control, it could be used in which the localization sensor fidelity is low, \eg urban areas, and big metropolitans. 
On the other hand, there are several limitations that should be overcome before moving to a real-life implementation of this proposal. 


Future work can focus on the other possibilities to assess the probability using other contextual information such as the age of the visible pedestrians, the possible actions engaged by the visible pedestrians (\eg presence of children playing soccer at the sidewalk, presence of distracted pedestrians due to use of cellphones or a conversation companion). 
\section*{ACKNOWLEDGMENT}
This work was partially funded by the United States Department of Transportation under award number 69A3551747111 for Mobility21: the National University Transportation Center for Improving Mobility. Any findings, conclusions, or recommendations expressed herein are those of the authors and do not necessarily reflect the views of the United States Department of Transportation, Carnegie Mellon University, or The Ohio State University. 

Additionally, the first author acknowledges financial support from Turkish Fulbright commission.






\bibliographystyle{IEEEtran}
\bibliography{bibfile}

\end{document}